\documentclass{article}
\usepackage{amsmath}
\usepackage{spconf}
\usepackage{hyperref}
\usepackage{enumitem}
\usepackage{dsfont}
\usepackage{comment}

\usepackage{graphicx} 
\usepackage{amssymb} 
\makeatletter 
\newcommand*{\rom}[1]{\expandafter\@slowromancap\romannumeral #1@} 
\setlist{nosep, leftmargin=14pt}
\ninept
\usepackage{mwe} 

\title{MOAB: Multi-modal Outer Arithmetic Block for fusion of histopathological images and genetic data for brain tumor grading}

\name{Omnia Alwazzan$^{\star \dagger}$ , Abbas Khan$^{\star \dagger}$, Ioannis Patras$^{\star \dagger}$ , Gregory Slabaugh$^{\star \dagger}$}
\address{$^{\star}$School of Electronic Engineering and Computer Science, Queen Mary University of London, UK\\$^{\dagger}$Queen Mary's Digital Environment Research Institute (DERI), London, UK}

\begin{document}
%
\maketitle
\begin{abstract}
Brain tumors are an abnormal growth of cells in the brain. They can be classified into distinct grades based on their growth. Often grading is performed based on a histological image and is one of the most significant predictors of a patient's prognosis; the higher the grade, the more aggressive the tumor. Correct diagnosis of the tumor’s grade remains challenging. Though histopathological grading has been shown to be prognostic, results are subject to interobserver variability, even among experienced pathologists. Recently, the World Health Organization reported that advances in molecular genetics have led to improvements in tumor classification. This paper seeks to integrate histological images and genetic data for improved computer-aided diagnosis. We propose a novel Multi-modal Outer Arithmetic Block (MOAB) based on arithmetic operations to combine latent representations of the different modalities for predicting the tumor grade (Grade \rom{2}, \rom{3} and \rom{4}). Extensive experiments evaluate the effectiveness of our approach. By applying MOAB to The Cancer Genome Atlas (TCGA) glioma dataset, we show that it can improve separation between similar classes  (Grade \rom{2} and \rom{3}) and outperform prior state-of-the-art grade classification techniques.

\end{abstract}
\section{Introduction}\label{sec:intro}

With the advent of artificial neural networks, many advanced medical imaging algorithms have been proposed to analyze histopathological cancer images for grade classification \cite{niazi2016visually,wan2017automated} and survival prediction \cite{tabibu2019pan,huang2019salmon}. Digitized histopathology provides whole slide images (WSI) and related phenotypic information. Literature suggests both single-modality \cite{wetstein2022deep} and multi-modality \cite{chen2020pathomic} methods for histopathological image analysis. However, approaches based on a single modality are limited and unable to exploit the complementary information present in other modalities for cancer prognosis prediction \cite{arya2020multi,yan2021richer}. This provides an opportunity to  integrate multi-modal data for more precise diagnosis.

A recent study \cite{braman2021deep} integrated 'omic data with image modalities using attention gating and a combination of concatenation and arithmetic operation-based fusion approaches. Braman et al. \cite{braman2021deep} presented an intermediate fusion deep multi-modality network that integrated radiology, pathology, genetics, and clinical data to predict glioma patients. Deep learning models were used in \cite{braman2021deep} to extract features from each modality and generate its corresponding embedding layer. Using the attention gated mechanism and an outer product operator, the authors were able to fuse significant features to improve prediction. 

A multi-modal fusion method was proposed by \cite{tan2022multi} to learn the combined feature representations of histopathological images and mRNA expression, through ResNet-152 and a sparse graph convolutional network respectively. The features were merged using a shared multilayer perceptron to predict survival analysis and cancer grade. Simailarly, Pathomic Fusion \cite{chen2020pathomic}, an integrative image-omic analysis method used a gated attention mechanism and a Kronecker product to combine features of different modalities. The image features were extracted using a combination of convolutional and graph convolutional neural networks, and a feed-forward network was trained separately to learn genomics features. Mobadersanya et al. \cite{mobadersany2018predicting} proposed survival convolutional neural networks (SCNNs) to merge a histology image and genomic biomarkers. 

\begin{figure*}[ht]
\centering
\includegraphics[scale=0.5]{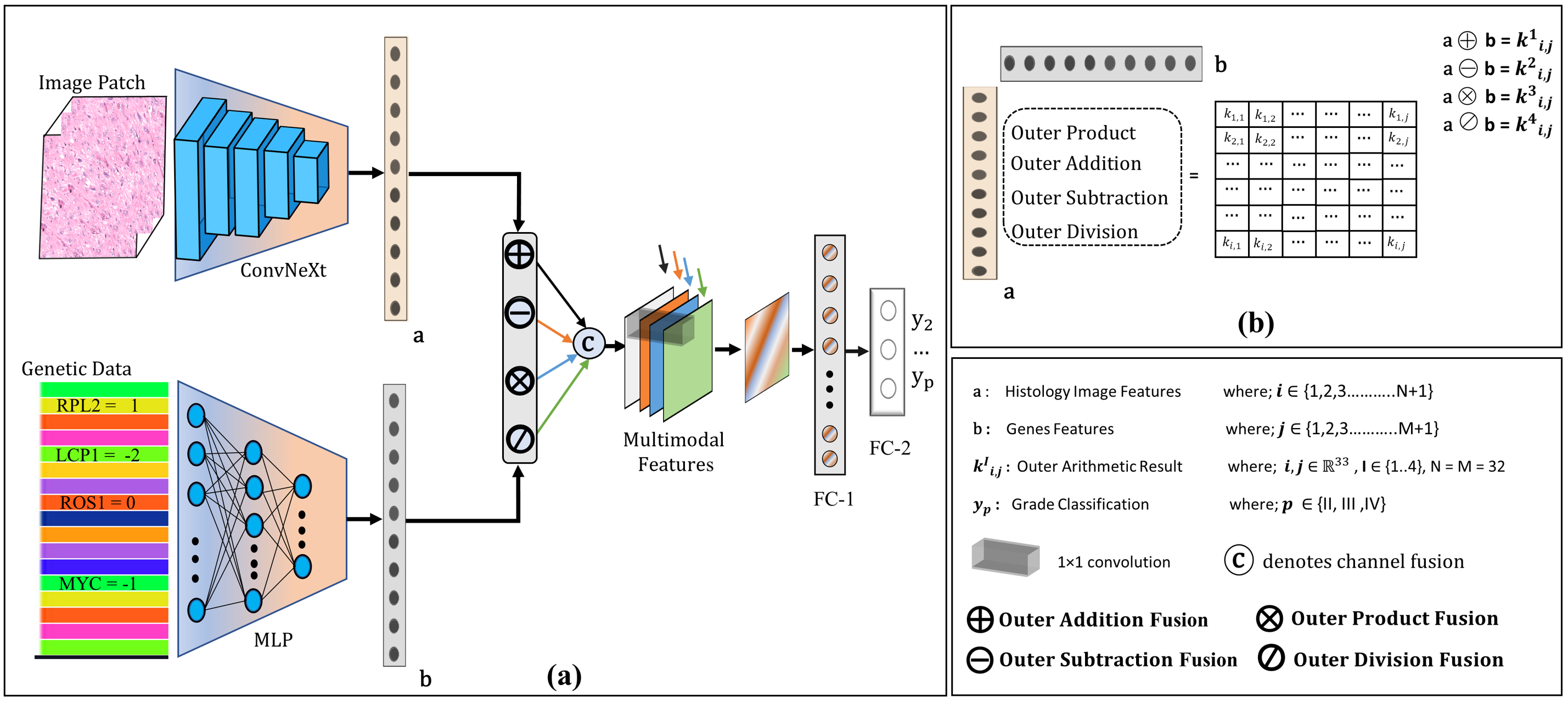}
\caption{Architecture of our proposed approach Multi-modal Outer Arithmetic Block (MOAB) fusion model. (a) MOAB (b) Generic outer arithmetic approach.}
\label{fig:fig1}
\end{figure*}

We note that previous work \cite{chen2020pathomic,tan2022multi} focused on Grade \rom{4} and did not consider Grade \rom{2} and \rom{3}. These two grades are minorities in the dataset compared to Grade \rom{4} and their similarity hinders typical classification techniques from detecting them accurately. However, Grade \rom{3} is considered malignant and as invasive as Grade \rom{4} \cite{hanif2017glioblastoma} which could make its classification as crucial as Grade \rom{4}. Hence, in this study, we show how to integrate rich features from genetic data and histology images to improve grading of tumors. 

We propose a novel fusion method that captures the interrelated features between disparate modalities for improved classification. An overview of our approach is presented in Fig.~\ref{fig:fig1}. Our paper makes the following contributions:
\begin{itemize}
    \item We introduce a novel Multi-modal Outer Arithmetic Block (MOAB) that fuses latent representations of different modalities using arithmetic operations. 
    \item We demonstrate MOAB’s ability to capture rich representations of fused data in a brain cancer application using a histology image and genetic data. Our method outperforms the previous state-of-art, by producing more discriminative fused features for classification. This is also demonstrated with t-SNE plots which visualize the improved separability of classes thanks to MOAB.
    \item Source code for all proposed fusion methodologies are available at \href{https://github.com/omniaalwazzan/MOAB}{https://github.com/omniaalwazzan/MOAB}
\end{itemize}

\section{METHOD}\label{sec:method}
From paired histology and genetic data with known cancer grades, we aim to learn and combine informative features that outperform single-modality classifiers in supervised learning. For this purpose, we propose a novel method for effective integration of multi-modal data.

We first provide a brief overview of the single modality backbone that we use, followed by in-depth explanations of the architecture of two components: outer arithmetic fusion methodologies and our proposed MOAB.

\subsection{Single modality classifiers} 
We initially created a single modality classifier for each data type (i.e. histology image and genetic data). 
For the histology images, a pre-trained ConvNeXt initialized with ImageNet weights was used. ConvNeXt is a recently proposed~\cite{liu2022convnet}, state-of-the-art CNN demonstrating how convolutional backbones can outperform transformers in image classification \cite{han2022convunext}. We replace the LayerNorm2D of the ConvNeXt classifier layer with a dropout = 0.2 followed by a fully connected (FC) layer that outputs three values representing the brain tumor grades (\rom{2}, \rom{3} and \rom{4}). We fine-tune ConvNeXt on the histology images. We observe that ConvNeXt is more robust than the VGG19 backbone implemented in \cite{chen2020pathomic} in terms of generalizing well to the histology images after a few iterations. 

A 3 consecutive blocks of FC layers with dimensions [80, 40, 32] 
 forming our multi-layer perceptron (MLP) was utilized to classify the genomic data. Each layer in the MLP is followed by a rectified linear unit (ReLU) activation and a layer normalization. After the second and third layers, a gentle dropout of 0.2 was applied. According to the results presented in Table~\ref{tab1}, the MLP achieves comparable performance to the ConvNeXt applied to the histological images.

\subsection{Outer Arithmetic Operations} 
Our work is based on Outer Product Fusion (OPF), which has been used previously \cite{braman2021deep} to combine feature vectors of different modalities. Consider two feature vectors $\mathbf{a} \in \mathbb{R}^{N\times1}$ and $\mathbf{b} \in \mathbb{R}^{M\times1}$. We first append a 1 to each feature vector, i.e. $\mathbf{a}_1 = [1; \mathbf{a}]$ and $\mathbf{b}_1 = [1; \mathbf{b}]$.  The OPF is expressed as
\begin{equation}
    (\mathbf{a}_1 \otimes \mathbf{b}_1)_{ij} = a_{1i}*b_{1j}, 
\end{equation}
for $i \in [1 ... N+1]$ and $j \in [1 ... M+1]$ resulting in an $(N+1) \times (M+1)$ matrix combining all pairs of features.  The appended 1 in both $\mathbf{a}_1$ and $\mathbf{b}_1$ ensures the original $\mathbf{a}$ and $\mathbf{b}$ vectors appear in the outer product matrix.

We extend the OPF by proposing the use of three new additional arithmetic operations: \emph{Outer Addition Fusion (OAF)}, \emph{Outer Subtraction Fusion (OSF)} and \emph{Outer Division Fusion (ODF)}.  The ODF is expressed as: 
\begin{equation}
    (\mathbf{a}_1 \oslash \mathbf{b}_1)_{ij} = a_{1i}/(b_{1j} + \epsilon), 
\end{equation}
where $\epsilon$ is a small number to avoid division by zero. 

For the outer addition and outer subtraction fusion, we append a 0 to each feature vector, i.e. $\mathbf{a}_0 = [0; \mathbf{a}]$ and $\mathbf{b}_0 = [0; \mathbf{b}]$. Then, the OAF is defined as
\begin{equation}
    (\mathbf{a}_0 \oplus \mathbf{b}_0)_{ij} = a_{0i} + b_{0j}, 
\end{equation}
and OSF as 
\begin{equation}
    (\mathbf{a}_0 \ominus \mathbf{b}_0)_{ij} = a_{0i} - b_{0j}, 
\end{equation}
As before, the appended 0 in both $\mathbf{a}_0$ and $\mathbf{b}_0$ ensures the original $\mathbf{a}$ and $\mathbf{b}$ vectors appear in the outer addition and subtraction matrices.
Each of the four outer arithmetic fusion operators will produce a matrix of size $(N+1) \times (M+1)$.

\subsection{Multi-modal Outer Arithmetic Block Fusion} \label{sec:FOA}
The MOAB fusion model uses the previously explained outer arithmetic fusion operations to capture essential features from the genomic and histology data. We perform the fusion on a four branch scheme. First, we obtain latent representation feature vectors $\mathbf{a}$ and $\mathbf{b}$ from histology and genomic data, respectively. As illustrated in Fig~\ref{fig:fig1}(a), $\mathbf{a}$ was extracted from the customized ConvNeXt network used previously for histology image classification. Inspired by \cite{chen2020pathomic}, we obtained 32 features by modifying the number of the output dimension of the last FC layer from the ConvNeXt. Similarly, 32 features were extracted from the last 
FC layer from our MLP to represent $\mathbf{b}$. 

Next, each branch will take $\mathbf{a}$ and $\mathbf{b}$, perform the four arithmetic operations to produce four multi-modal feature maps, $A_{i,j}$, $S_{i,j}$, $P_{i,j}$, and $D_{i,j}$, that represent the OAF, OSF, OPF and ODF respectively. Moreover, we utilized a sigmoid function to compress our feature maps between [0,1]. The sigmoid is employed due to the addition of the $\epsilon$ = 1.2e\textminus20  avoiding zero division in the ODF branch. The sigmoid was applied to all four branches to maintain compatible fusion between all branches. 

The four matrices $A_{i,j}$, $S_{i,j}$, $P_{i,j}$, and $D_{i,j}$, are concatenated along the channel dimension into a multi-modal tensor $M_{ij} \in \mathbb{R}^{4\times33\times33}$. Then, a 2D convolution layer is applied with a filter $f=1$ and a stride $s=1$ to take advantage of interrelated interactions and produce one condensed multi-modal feature tensor $M_{ij}^* \in \mathbb{R}^{1\times33\times33}$. We hypothesize that channel fusion will maintain the proximity of closer points and will use fewer parameters compared to a typical concatenation. By combining features across the channel dimension, we massively decrease the dimension of the final FC layer from $4356$ to $1089$ in terms of concatenation. It has been noted that $f={3,5}$ leads to a decrease in the model’s performance. Hence, we avoid using filter $f>1$.

Finally, $M_{ij}^*$ is passed to two FC layers to perform the final classification. A mild dropout$=$0.1 is employed before the final layer to avoid overfitting. We also adopt the orthogonal weight initialization introduced in \cite{chen2020pathomic} for the MLP component in the proposed architecture.



\section{EXPERIMENTS AND RESULTS}
\label{sec:exp}

\begin{figure*}[ht]
\centering
\includegraphics[scale=0.6]{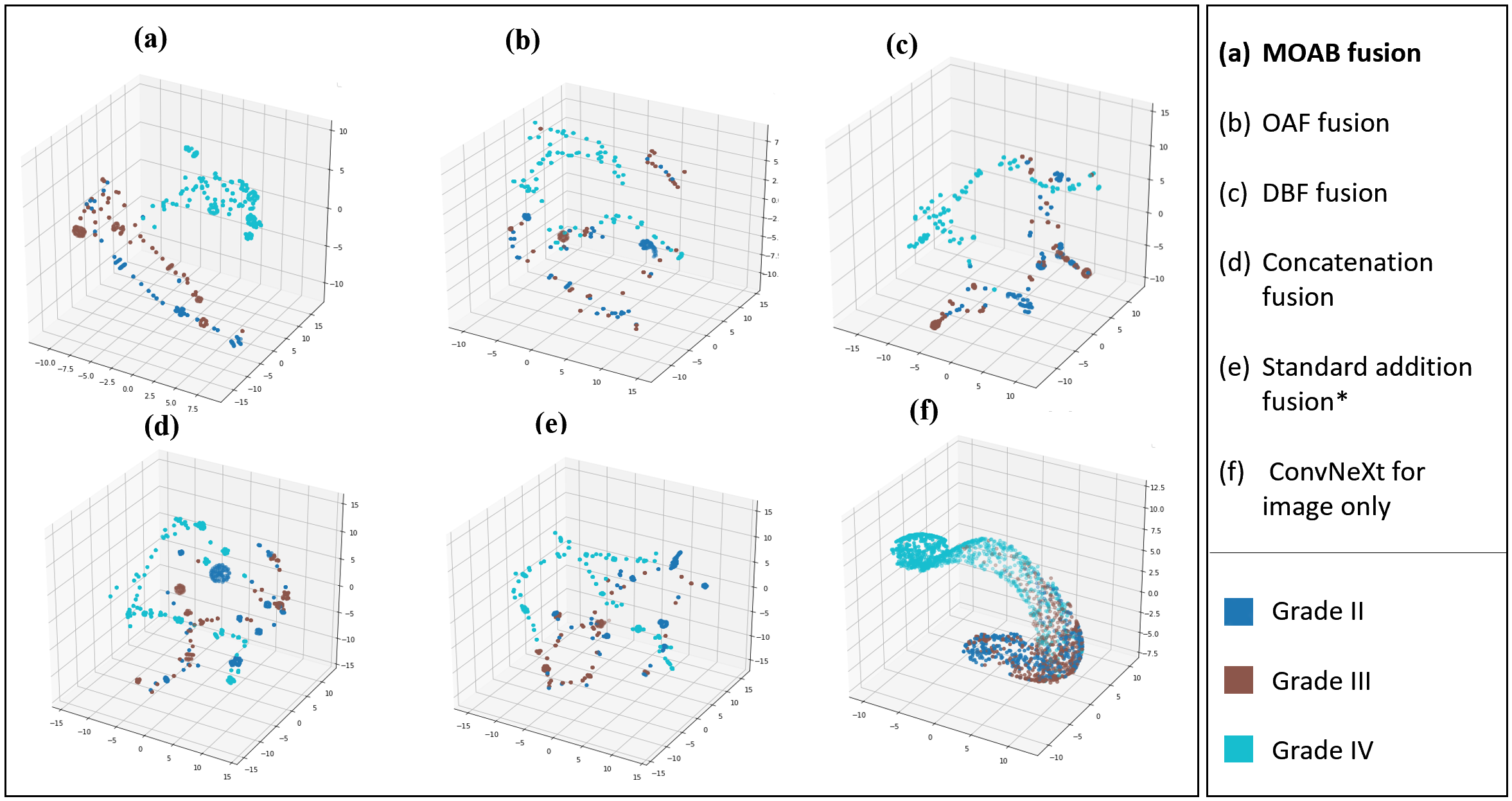}
\caption{t-Distributed Stochastic Neighbor Embedding (t-SNE) visualization for fusion models. Note in (a), the MOAB fusion provides better separability between brain tumor grades.}.
\label{fig:fig2}
\end{figure*}

\subsection{Dataset} \label{sec:dataset}
The glioma dataset was preprocessed as proposed by Chen et al.~\cite{chen2020pathomic}, and provides paired histology images and gene expression derived from The Cancer Genome Atlas (TCGA) repository. The TCGA is a cancer database containing paired high-throughput genome diagnostic whole slide images with ground-truth histological grade labelling. There are 769 patients belong to 1505 histological region-of-interests (ROIs) for astrocytomas and glioblastomas in the TCGA-GBMLGG project. The selection of ROI derived from histopathological whole slide images (WSIs)  was generated, revised, and reviewed by \cite{mobadersany2018predicting}. Each patient had 1-3 ROIs from diagnostic WSI. For the genomic data, 80 features include 79 copy number variations (CNVs) and one mutation status. Similar to \cite{chen2020pathomic}, we removed cases with missing grades, resulting in 396, 408, and 654 ROIs for Grade \rom{2}, \rom{3} and \rom{4}, respectively, and corresponding to 736 cases. 
As in \cite{chen2020pathomic}, ROIs are used for training models, while 9 overlapping patches extracted per each ROI derived from the testing cohort were used for testing. Each image was regarded as a single data point as in \cite{chen2020pathomic}, with genetic and ground-truth label data copied over.

\subsection{Implementation Details}   \label{sec:Implement}
Proposed algorithms were implemented in PyTorch using the Adam optimizer and a cross-entropy loss function for all experiments. For the single modalities, we empirically found the best learning rate of 0.001. While for fusion models, to avoid overfitting, we regularized our networks with a weight decay of size 0.0005 and a 0.005 learning rate. A batch size of 8 was set for all experiments. All models were trained for 10 epochs. The parameters for ConvNeXt, MLP and MOAB are 87M, 11K and 88M respectively. A cluster of NVidia A100 GPUs was utilised for all experiments.
\subsection{Ablation Study} \label{sec:abation}
As shown in Table~\ref{tab1}, we validate the performance of MOAB fusion through multiple ablation studies. In an identical Monte Carlo 15-fold cross-validation train-test split obtained from \cite{chen2020pathomic, mobadersany2018predicting}, our ablation studies compare different model configurations and fusion methodologies. In addition, as highlighted in Fig.~\ref{fig:fig2}, a (t-SNE) algorithm \cite{van2008visualizing} was applied to the inference split for ablated fusion models to visualize the high-dimensional relationships between the three learned grades (\rom{2}, \rom{3} and \rom{4}) in three-dimensional space. In t-SNE plots, nearby dots represent comparable samples, whereas distant dots reflect dissimilar ones. We empirically set an optimal perplexity and a max iteration to 80 and 1000. 


\begin{table}[ht!]
\caption{Ablation studies.}\label{tab1}
    \centering
    \begin{tabular}{l|c|c|c}
    \hline
         \ Method &  F1 (Grade IV) & F1-Micro & F1-Macro\\ \hline
        CNN (Image) & 0.888 & 0.715 & 0.586 \\ 
        MLP (Genes) & 0.901  & 0.700 & 0.626  \\ 
        Concatenation & 0.910 & 0.726 & 0.658\\ 
        OAF  & 0.944 & 0.740 & 0.675 \\ 
        DBF  & 0.916 & 0.727 & 0.660  \\ 
        Standard Addition* & 0.933  & 0.730 & 0.665 \\ \hline
        \shortstack[l]{MOAB\\(VGG19\textunderscore bn+MLP))} & 0.945& 0.752 &0.689 \\ \hline
        \shortstack[l]{\textbf{MOAB}\\ \textbf{(ConvNext+MLP)}}  & \textbf{0.956} &\textbf{0.766} & \textbf{0.697}\\ \hline
    \end{tabular}
\end{table}

At first, single classifiers for both gene and image data were constructed to draw a fair comparison to multi-modality fusion. Next, we show results for a simple concatenation fusion model to further investigate the ability of other fusion methodologies. To explore the capability of MOAB, we produced two ablated fusion models: the Outer Addition Fusion (OAF) and a Dual-branch channel Fusion (DBF). Note that OAF has no channel fusion, features are combined as shown in Fig.~\ref{fig:fig1}(b) performing $A_{ij} \in \mathbb{R}^{33\times33}$ and followed by FC layers similar to MOAB. However, DBF imitates the same configuration as MOAB but with two branches: OAF and OPF. From Table~\ref{tab1} and Fig.~\ref{fig:fig2}(b) we observe that OAF performs better than the DBF even with a simple integration. This might occur due to the redundant features obtained from OAF, and OPF. Yet, DBF achieves similar metric scores (F1 and F1-Micro) to that of concatenation, the separated cluster shown in Fig.~\ref{fig:fig2}(c) is slightly better than those of Fig.~\ref{fig:fig2}(d), which might reflect the effect of OAF operation on capturing correlated features.

As demonstrated in Table~\ref{tab1}, the best results were obtained utilising MOAB. This is further supported by Fig.~\ref{fig:fig2}(a), which shows MOAB fusion has the most separated classes among all other ablated fusion models. By comparing Fig.~\ref{fig:fig2}(a) with Fig.~\ref{fig:fig2}(f) we can clearly inspect the positive effect of integrating features from genetic data and histology images. Table~\ref{tab1} summarizes that utilizing all four branches in the channel fusion has significant improvements on the classifier performance namely F1-Micro and F1-Macro. F1-micro score is calculated from the micro-averaged precision (miPr), and a micro-averaged recall (miRe) and defined as the following:

\begin{equation}
    \begin{aligned}\label{eq:avfF1}
            miPr &= \frac{\sum_{i=1}^{r} TP_i}{\sum_{i=1}^{r} (TP_i + FP_i)} \\ 
            miRe &= \frac{\sum_{i=1}^{r} TP_i}{\sum_{i=1}^{r} (TP_i + FN_i)}
     \end{aligned}
\end{equation}
The $TP_i$, $FP_i$ $FN_i$ denote true positive, false positive and false negative rates at a cell $i$ in the confusion matrix. 
Then the  micro-averaged F1 score (miF1) is defined as the mean of $miPr$, $miRe$ values.

We use the F1-micro score as the primary metric to evaluate our results, following  \cite{tan2022multi},\cite{chen2020pathomic}, as it is an appropriate evaluation metric for imbalanced class data. Notably, F1-Grade \rom{4} is an easier classification task given that it dominates the majority of our dataset; still, we include results for comparison purposes. Furthermore, to quantify the performance of MOAB, we calculate the F1-macro average which treats classes equally and gives a sense of MOAB's effectiveness on minority classes. The macro averaging method is more challenging for our task because it is more strongly influenced by the performance of rare classes \cite{yang1999re}. Table~\ref{tab1} proves that MOAB manages to achieve strong performance.
Finally, to conclusively say OAF is preserving the interlarded features between modalities, we conduct an experiment for a standard addition fusion model shown in Table~\ref{tab1}, where we increase the parameters of the standard addition (denoted as an *) to have roughly the same learnable parameters as the OAF. Results of the standard addition exhibited in Table~\ref{tab1} and Fig.~\ref{fig:fig2}(e) indicate that OAF outperforms standard addition fusion*.


We have also switched the backbone of the MOAB from ConvNxet to the VGG19 with batch normalization used in \cite{chen2020pathomic}, again, our obtained results outperform the Pathomic fusion \cite{chen2020pathomic} approach. 
We conclude that MOAB is improving the classifier thoroughly compared to all other configurations presented in Fig.~\ref{fig:fig2}. We observed that MOAB is simple to implement and extracts rich interaction combining modalities.


\subsection{Comparison Experiments}\label{sec:compare}
We compared our method with the Pathomic fusion \cite{chen2020pathomic} and a MultiCoFusion \cite{tan2022multi} on the glioma dataset. The results are shown in Table~\ref{tab2}. The TCGA glioma dataset has been used with 15-folds in both \cite{chen2020pathomic} and~\cite{tan2022multi}, however, the approach used in \cite{tan2022multi} is slightly different than ours, in which they additionally use mRNA expression data using a graph CNN. 
\begin{table}[ht!]
\caption{Comparison experiments.}\label{tab2}
    \centering
    \begin{tabular}{l|c|c}
    \hline
         \ Model & F1 (Grade IV) &  F1-Micro \\ \hline
        Pathomic SNN \cite{chen2020pathomic}& 0.857 $\pm$ 0.017 & 0.652 $\pm$ 0.015 \\ 
        Pathomic (CNN+SNN) \cite{chen2020pathomic}& 0.913 $\pm$ 0.011   & 0.730 $\pm$ 0.019\\ 
        MultiCoFusion \cite{tan2022multi} & 0.998 $\pm$ 0.005 & 0.759 $\pm$ 0.032\\ \hline
        MLP (Genes) & 0.901 $\pm$ 0.051  & 0.700 $\pm$ 0.042 \\ 
        \shortstack[l]{MOAB\\(VGG19\textunderscore bn+MLP))} & 0.945 $\pm$ 0.012& 0.752 $\pm$ 0.002 \\ 
        \shortstack[l]{\textbf{MOAB}\\ \textbf{(ConvNext+MLP)}}  &0.956 $\pm$ 0.000  &\textbf{0.766 $\pm$ 0.001}\\ \hline
    \end{tabular}
\end{table}
According to the ablation studies shown in Table~\ref{tab1} and the comparison presented in Table~\ref{tab2}, the F1-Micro of MOAB is superior to that of the other models. Experiments were performed 15 times; the results shown in the tables represent the mean of the total of all evaluation metrics for 15-folds, as demonstrated in \cite{chen2020pathomic} and \cite{tan2022multi}.

\section{CONCLUSION}
\label{sec:conc}
In this paper, we proposed a novel Multi-modal Outer Arithmetic Block (MOAB) that fuses features from histology and genetic data. Involving features from all arithmetic operations encourages the classifier to capture intrinsic relations between modalities, ultimately enhancing the overall prediction. The channel fusion strategy enables MOAB to reduce the parameters of conventional fusion methods and enhances the classifier's knowledge. All conducted experiments demonstrate the effectiveness of proposed approach. Last, we note that the architecture of MOAB is transferrable to other multi-modal problems as it can combine modalities of any type with any CNN backbones. In the future, We plan to experiment MOAB with additional datasets and expand it to integrate more modalities.       



\section{Acknowledgments}
The authors appreciate the support of the University of Jeddah and the Saudi Arabia Cultural Bureau.  This paper utilised Queen Mary's Andrena HPC facility. This work also acknowledges the support of the National Institute for Health and Care Research Barts Biomedical Research Centre (NIHR203330), a delivery partnership of Barts Health NHS Trust, Queen Mary University of London, St George’s University Hospitals NHS Foundation Trust and St George’s University of London.
\section{COMPLIANCE WITH ETHICAL STANDARDS}
The licence related to the open-access data ensured no ethical approval was necessary.

\bibliographystyle{IEEEbib.bst}

\bibliography{refs.bib}

\end{document}